\documentclass[a4paper, 10 pt, conference]{ieeeconf}  

\IEEEoverridecommandlockouts                              

\usepackage{graphics} 
\usepackage{epsfig} 
\usepackage{amsmath} 

\usepackage{csquotes}
\usepackage{comment}
\usepackage{subcaption}
\usepackage{graphicx} 
\usepackage{mathtools}
\usepackage{cite}

\makeatletter
\let\NAT@parse\undefined
\makeatother
\usepackage[breaklinks=true,letterpaper=true,colorlinks,bookmarks=false]{hyperref}

\title{\LARGE \bf
Initiation of Interaction Detection Framework using a Nonverbal Cue for Human-Robot Interaction
}

\author{Guhnoo Yun$^{1}$, Juhan Yoo$^{2}$, Kijung Kim$^{1}$ and Dong Hwan Kim$^{1}$
\thanks{$^{1}$Guhnoo Yun and Dong Hwan Kim are with Korea Institute of Science and Technology, Seoul, 02792, Republic of Korea
        {\tt\small \{doranlyong, plan100day, gregorykim\}@kist.re.kr}}%
\thanks{$^{2}$Juhan Yoo is with the Department of Computer Science, Semyung University,
        Jecheon-si, 27136, Republic of Korea
        {\tt\small unchinto@semyung.ac.kr}}%
}

\begin{document}

\maketitle
\thispagestyle{empty}
\pagestyle{empty}

\begin{abstract}

This paper describes an initiation of interaction(IoI) detection framework without keywords for human-robot interaction(HRI) based on audio and vision sensor fusion in a domestic environment. In the proposed framework, the robot has its own audio and vision sensors, and can employ external vision sensor for stable human detection and tracking. When the user starts to speak while looking at the robot, the robot can localize his or her position by its sound source localization together with human tracking information. Then the robot can detect the IoI if it perceives the face of the speaker faces the robot. In case that the user does not speak directly, the robot can also detect the IoI if he or she looks at the robot for more than predefined periods of time. A state transition model for the proposed IoI detection framework is designed and verified by experiments with a mobile robot. In order to implement and associate our model in a robot architecture, all the components are implemented and integrated in the Robot Operating System(ROS) environment.

\end{abstract}

\section{INTRODUCTION}
\label{sec:introdution}
There has been growing interest in the field of socially assistive robots~\cite{broekens2009assistive, mollaret2016multi, fardeau2023impact}.
The goal of socially assistive robots is to provide assistance to human users through social interaction~\cite{benedictis2023dichotomic, FeilSeifer2005}.
Especially, it becomes essential for the elderly who live alone or suffer from mild cognitive impairment or Alzheimer's disease.
Various problems from cognitive decline lead them to experience stress and loss of confidence, and also to become irritable, which put them at health risk especially in considering their frailty~\cite{mollaret2016multi, trick2019multimodal}.
Generally, it takes huge amount of money and manpower to care for them, and therefore a care service using a socially assistive robot can be a relevant solution to these kinds of social problems~\cite{broekens2009assistive, vcaic2018service, shareef2023machine}.

For a socially assistive robot, robot services including a care service should be performed in a socially interactive manner. During the interaction, the robot should automatically detect who needs a service and decide when the service be provided. It is very important for a robot to know if a user has an intention to interact with it, and automatic detection of it makes it possible for the robot to initiate a proper service protocol. Traditionally and practically, the initiation of interaction(IoI) detection has been achieved by detecting a special hotword such as a name of a robot or a service agent~\cite{alexa2023, chen2013continuous}.
In these methods, the user calls the name of the robot or the agent, and then the robot detects it and initiates interaction with the user. However, it would be a more efficient way to interact with the robot if the user doesn't need to say special keywords whenever he or she wants to ask for robot services.

In this paper, therefore, we propose a new framework for IoI detection based on audio and vision sensor fusion in a domestic environment.
The robot with the proposed framework can detect the IoI without using any of user's speech or a certain specific keyword.
The proposed framework detects the IoI by combining person tracking information from the vision sensor such as the Azure Kinect and the user's sound source localization information. With this information, our system perceives the IoI when a person looks at the robot for more than a predefined duration or talks while looking at the robot.

\begin{figure}[t]
\vspace{0.3cm}
\centering
    \includegraphics[width=0.9 \linewidth]{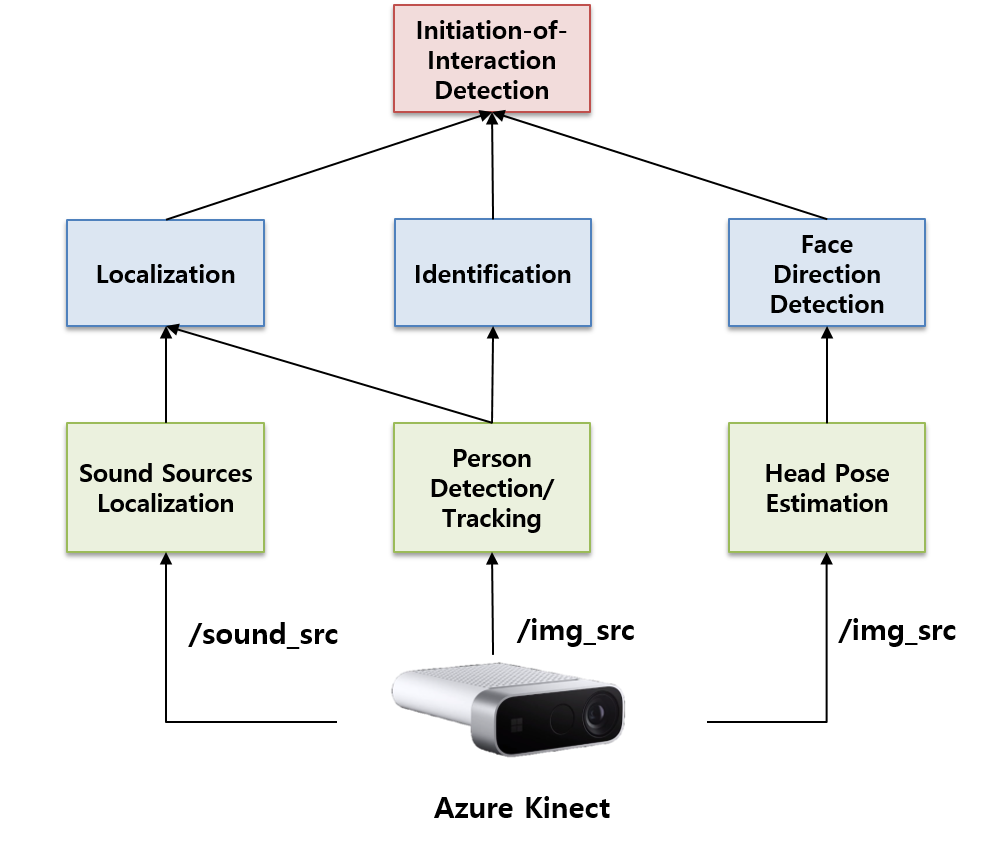}
    \caption{\textbf{The overview of the proposed IoI detection framework}. The blocks and arrows present ROS nodes and the data flows.}
    \label{frame_work}
\end{figure}

We use our previous monitoring robot \cite{kwon2023heterogeneous} as a robotic platform and install the Azure Kinect on the robot. This sensor is employed as a vision sensor and also includes an audio sensor since it contains a 7-microphone array. Our framework is developed on the Robot Operating System(ROS) ~\cite{quigley2009ros}. Fig~\ref{frame_work}. illustrates the overall structure of the proposed system. The rectangle blocks and arrows represent individual ROS nodes and the data flows between nodes, respectively.

The following sections outline the structure of this paper: Section ~\ref{sec:relatedworks} provides an overview of related works. Sections ~\ref{sec:perceptions} and ~\ref{sec:interaction} describe individual modules and present the proposed framework for the IoI detection, respectively. Section ~\ref{sec:experiment} discusses experimental results, and the paper concludes with a summary in Section ~\ref{sec:conclusions}.

\section{Related Works}
\label{sec:relatedworks}
Recently, research on the IoI detection has attracted considerable attention.
Generally, understanding user intentions for interaction plays a fundamental role in establishing coherent communication among people~\cite{blakemore2001perception, mollaret2016multi, trick2023can,bi2023human}.
The IoI is the situation before or just at the moment when both a human and a robot establish the common belief that they are sharing a conversation~\cite{carton2013proactively, shi2015measuring}.
According to~\cite{katagiri2006conversational}, people first need to establish a conversational space together with their conversational partners for interaction. The conversational space formation normally proceeds with conversational partners first approaching each other to form a spatial aggregate and then exchanging eye gazes and various forms of greetings.

Mollaret \emph{et al.}~\cite{mollaret2016multi} detected user's intention for interaction based on RGB-D and audio data fusion. They only considered situations in which user initiates the interaction by talking. That is, this method cannot detect the IoI when the user remains silent. In~\cite{shi2015measuring} and~\cite{shi2011spatial}, markers were attached to the human body to detect precise position and spatial formation. However, users may find it burdensome to continuously affix numerous markers to their bodies throughout the day.

Carton \emph{et al.}~\cite{carton2013proactively} and Truong \emph{et al.}~\cite{truong2017approach} present a trajectory planning method enabling autonomous robots to proactively approach people in dynamic environments to initiate a conversation. However, these methods are not appropriate for the care service robot in home environment. For instance, The care robot for the elderly must notice the user's intention of interaction before approaching the user.

Also, Ooko~\emph{et al.}~\cite{ooko2011estimating} have proposed a method based on head pose data. First, they analyzed the correlation between head pose information and a user's conversation, and then established an estimation model by applying a decision tree learning algorithm. 

In recent studies, multimodal features are employed to detect a human's intention for interaction in diverse human-robot interaction settings. For instance, Trick ~\emph{et al.} ~\cite{trick2023can} used features like head orientation, shoulder orientation, distance, speech activity recognition, and hotword detection to train multimodal probabilistic classifiers. However, this approach requires robots should always observe and record a user's behaviors, distance, and speech activities.


In this paper, we propose a simple approach for detecting whether a user intends to engage in an interaction without any hotword detection.

\section{Individual Perception Components}
\label{sec:perceptions}
In this section, we introduce a ROS framework for IoI detection through the fusion of audio and vision sensors.  It incorporates essential perception components to process data from these sensors.

\subsection{Sound Source Localization}
We utilize the open-source robot audition software HARK (Honda Research Institute Japan Audition for Robots with Kyoto University) for the localization of sound sources in user speech. \cite{hark, roshark}. It consists of sound source localization modules, sound source separation modules and automatic speech recognition modules of separated speech signals that works on any robot with any microphone configuration. We applied a sound source localization module of HARK to detect the direction of the user's speech sound. The sound source localization of HARK is based on the MUliple SIgnal Classification(MUSIC) method \cite{schmidt1986multiple}, which is one of the most successful methods to solve direction-of-arrival(DOA) problems \cite{iordache2013music, devaney2005time, ciuonzo2015performance}. The MUSIC method localizes sound sources based on source position and impulse responses (transfer function) between each microphone. In our integration, the sound source localization component is combined with the human detection and tracking component to obtain a position of a user who speaks.

\subsection{Person Detection and Tracking}
The above-mentioned sound source localization method provides directional information about a speaking user but has limitations in determining their exact location. To determine the user's precise location, we combine person detection and tracking with sound source localization. In this paper, we employ YOLOv7 \cite{wang2023yolov7} and DeepSort \cite{wojke2017simple} to detect and track the user locations based on RGB data. YOLOv7 is a real-time single-stage object detection system that divides the input image into a grid. Each grid cell predicts bounding boxes and class probabilities for the objects it contains. DeepSORT, an extension of the SORT (Simple Online and Realtime Tracking) algorithm ~\cite{bewley2016simple}, complements YOLOv7 in tracking. After YOLOv7 identifies objects, DeepSORT assigns and maintains unique IDs for each detected person. It uses Kalman filtering for motion prediction and a deep association metric for appearance matching. This combination ensures robust tracking even during occlusions and varying poses, crucial for real-time applications.

\subsection{Localization}
The localization component decides the position of a user who speaks based on the person detection and tracking component and the sound source localization component. From the person detection and tracking component, persons in a domestic environment are detected and tracked independently. Then we find the best-matched person when a user starts to speak by using the information obtained from the sound source localization component. Let us assume that the robot is located at the origin, and denote ${\bf{X}}_i$ as a tracked position of the $i$th person and $\bf{S}$ as a direction vector of a speech sound source, respectively. The criterion to determine the position of a user who speaks is as follows:

\begin{equation}
\label{eq_localization}
\bf{\hat{X}} = \arg \max_{{\bf{X}}_i} <{\bf{X}}_i, \bf{S}>,
\end{equation}
where $<\,\, , \,>$ means the dot product of vectors. Even though a certain ${\bf{X}}_i$ satisfies the above criterion, if the angle between ${\bf{X}}_i$ and $\bf{S}$ is larger than a predefined threshold, $\delta_{l}$, it is rejected. The typical value for $\delta_{l}$ is $15$ degrees.

\subsection{Face Direction Detection with Head Pose Estimation}
\begin{figure}[t]
\vspace{0.3cm}
\centering
    \includegraphics[width=0.9 \linewidth]{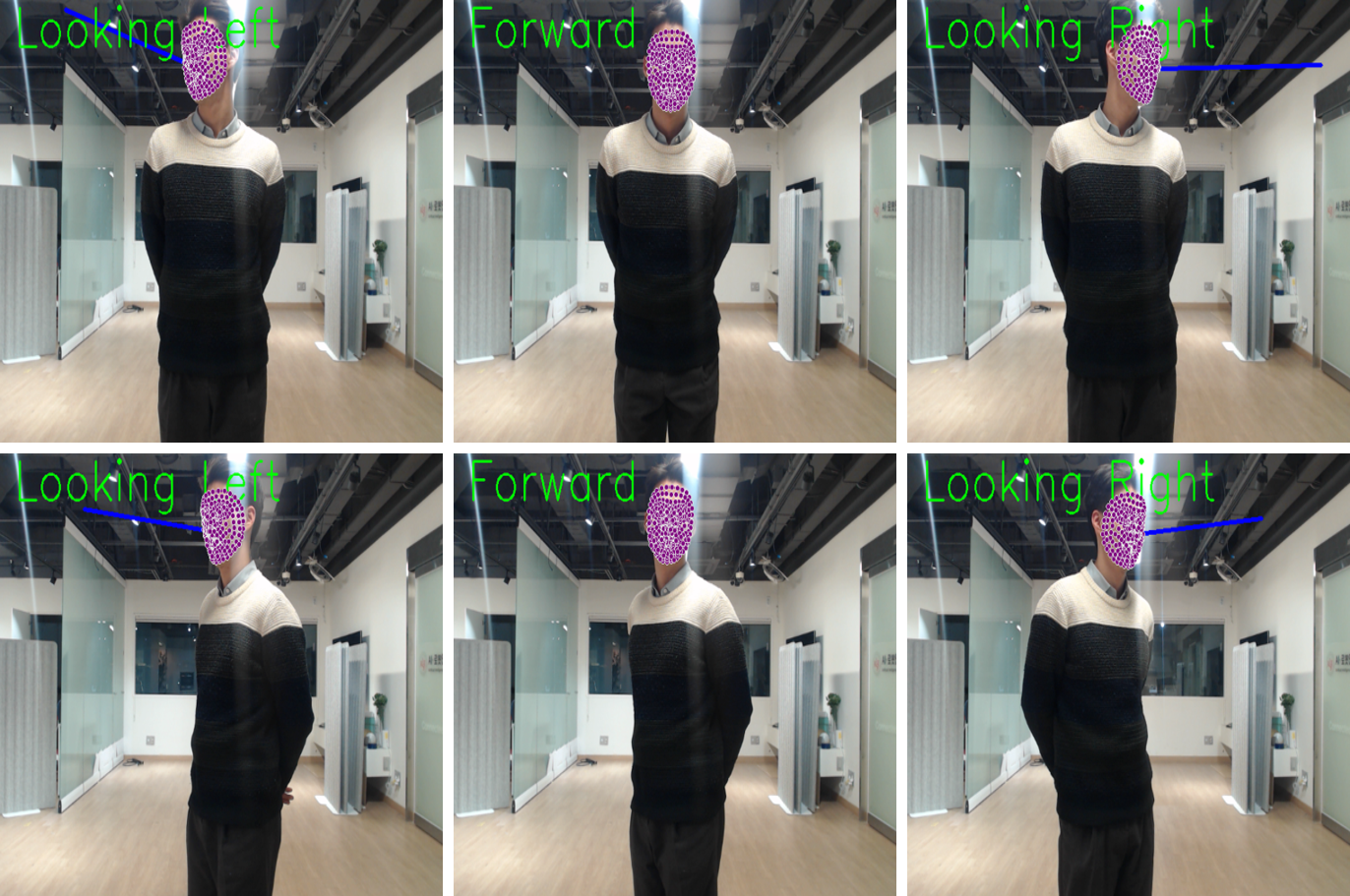}
    \caption{\textbf{Face direction detection with the head pose}. In the series of scenes depicted, each column represents the direction of gaze: looking to the left, front, and right, in that order.}
    \label{face_direction}
\end{figure}

After identifying the position of a speaking user, it is necessary to determine if he or she is directing their attention toward the robot to initiate an interaction. Generally, human interactions commence with eye gaze as a non-verbal cue for initiating conversation \cite{goffman2008behavior, argyle1972non, saran2018human, kiilavuori2021making}. However, discerning human gaze at a distance presents a significant challenge for robots in real-world settings, particularly without specialized devices such as glasses-type eye trackers or dedicated gaze-tracking equipment. Furthermore, the complexity and subtlety of human gaze direction mean that remote camera-based eye-gaze tracking is susceptible to various factors, including lighting and occlusions. To bypass this challenge, instead of detecting human eye gazes, we detect the frontal face with an assumption that the face direction of a user is toward the object he or she is looking at. Therefore, we infer that a user intends to engage in interaction with the robot if their head is oriented towards it. We simply implement face direction detection using the head pose estimation of MediaPipe \cite{lugaresi2019mediapipe}. MediaPipe is a versatile framework that offers a robust solution for the development and deployment of perception algorithms and models. It enables efficient detection through its pre-built model. Input frames are processeed using a pre-trained convolutional neural networks (CNNs) model that can efficiently identify facial features. Additionaly, the framework's CNNs are optimized for real-time, enabling face and landmark detection even on lower-powered devices. Its modular design allows for easy integration and adaptation, making it suitable for various applications from security to interactive media. The implementation results are shown in Fig \ref{face_direction}. In a more advanced IoI detection framework, however, this component is reserved for future enhancement, potentially through integration with a sophisticated eye gaze tracking-based method.

\begin{figure*}[!t]
	\centering
	\includegraphics[width=0.8 \linewidth]{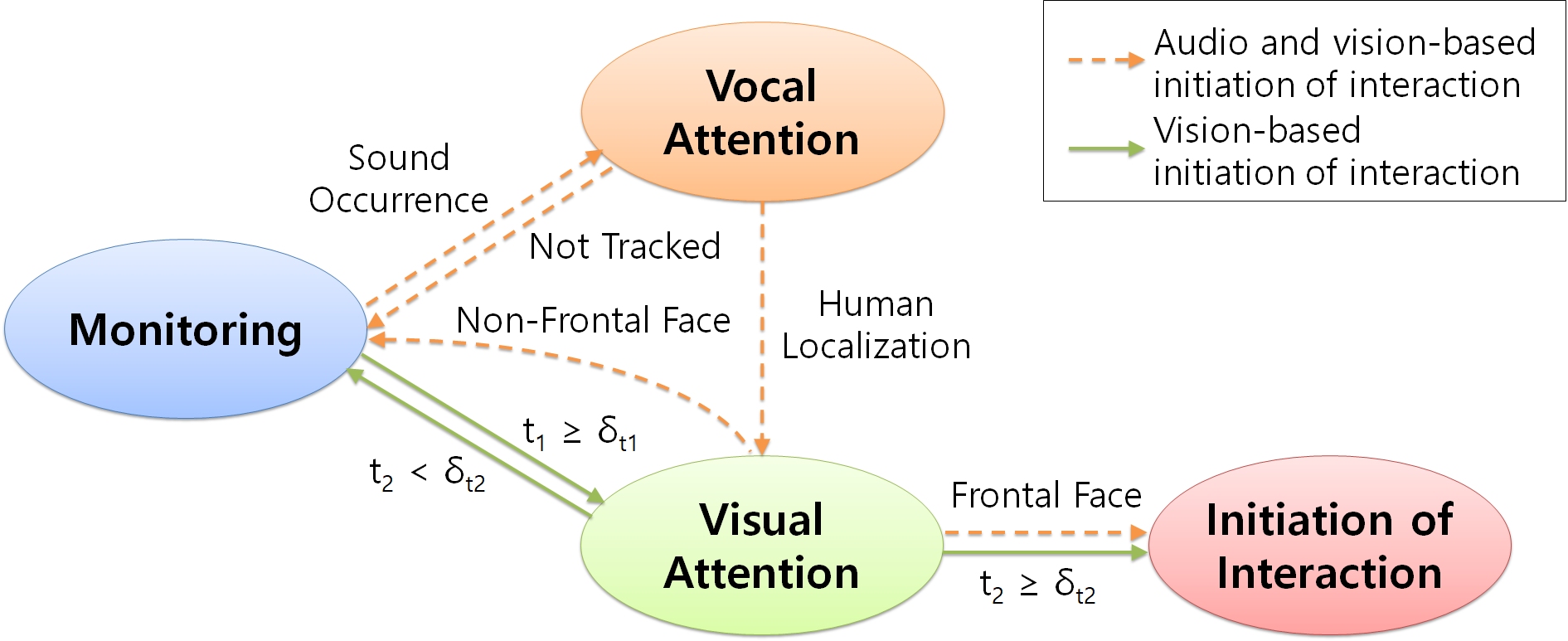}
	\caption{{\bf The proposed state transition model}. Initially, the robot keeps the monitoring state. According to the perceived data, the state can be changed to the vocal or visual attention state. Finally, the state becomes the IoI state when it meets criteria for it.}
	\label{fig_3}
\end{figure*}

\section{Initiation of interaction detection}
\label{sec:interaction}
In the proposed framework, IoI is identified either when a user initiates speech while facing the robot, or when the user's gaze is directed at the robot for a predefined duration. Both audio and visual data are utilized to ascertain the state of interaction with the robot. We design this framework with a state transition model as shown in Fig.~\ref{fig_3}. Initially, the robot is on the monitoring state. According to the perceived audio and visual sensor data, the robot's state can be changed by applying the proposed state transition model. In the proposed state transition model, the IoI state can be reached via two distinct paths. The first path involves simultaneous engagement in both vocal and visual attention states, occurring when the user begins speaking while concurrently looking at the robot. We call this first type of IoI detection {\it audio and vision-based IoI detection}. The other path is through the visual attention state alone, occurring when the user maintains their gaze on the robot for a predefined duration without speaking. We call this second type of IoI detection {\it vision-based IoI detection}.

\subsection{Audio and Vision-Based IoI Detection}
Audio and Vision-Based IoI detection occurs when the user starts to speak to the robot while looking at the robot. Initially, the robot monitors people in the environment with their tracking information obtained from the person detection and tracking component that uses vision sensors installed on the wall in the environment. When a user begins speaking, the robot utilizes its sound source localization component to detect the direction of the speech. Subsequently, the robot transitions to the vocal attention state, as indicated by the dashed line path in Fig.~\ref{fig_3}. If there is no tracked person whose face direction is toward the robot, the robot's state returns to the previous monitoring state. This backward transition may also occur in scenarios where no human is speaking, but a TV or radio is turned on. Otherwise, the localization component provides the user's position information, prompting the robot to transition to the visual attention state. If the face detection and face direction detection components fail to detect the frontal face of the user who speaks at the localized position, the robot's state also returns to the monitoring state. This backward transition may occur in scenarios where the user is speaking, but not to the robot, among other situations. If the frontal face is detected at the localized position, the robot transitions to the IoI state. The proposed audio and vision-based IoI detection is expressed as follows:

\begin{equation}
\label{eq_1}
z_{a}(s, f, h)=
\begin{cases}
1, & \mbox{for } s=1 \mbox{ , } f=1 \mbox{ and } h=1\\
0, & \mbox{otherwise},
\end{cases}	
\end{equation}
where $z_{a}$ represents an indicator function for the audio and vision-based IoI state. $s$ and $f$ mean a sound source and a frontal face of the user, respectively. $s=1$ and $f=1$ if sound source and frontal face are detected, respectively, and $s=0$ and $f=0$, otherwise. Similarly, $h=1$ if a user is tracked by the person detection and tracking component, and $h=0$, otherwise.

\subsection{Vision-Based IoI Detection}

In the vision-based IoI detection method, the robot identifies potential interaction when a user gazes at it for longer than a predetermined duration without speaking. Initially, the robot remains in the monitoring state. As mentioned in the audio and vision-based IoI detection, the robot monitors people with their tracking information. If the robot detects the frontal face of a tracked person, it begins to measure the duration of this face direction. Let us denote this time duration as $t_{1}$. The robot's state can be changed to the visual attention state along the solid line path shown in Fig.~\ref{fig_3} when the duration is greater than or equal to a predefined duration, $\delta_{t_{1}}$. This forward transition is expressed as follows:

\begin{equation}
\label{eq_2}
z_{v1}(t_{1})=
\begin{cases}
1, & \mbox{for } t_{1} \geq \delta_{t_{1}} \\
0, & \mbox{for } t_{1} < \delta_{t_{1}},
\end{cases}	
\end{equation}
where $z_{v1}$ represents an indicator function for the visual attention state.

Then, the robot's state can move to the IoI state only if the user is still facing the robot for more than another predefined duration. The robot measures another time duration, $t_{2}$ from when its state is changed to the visual attention state. If $t_{2}$ is greater than or equal to a predefined threshold, $\delta_{t_{2}}$, its state is changed to the IoI state. Otherwise, the robot state returns to the initial monitoring state. The forward transition is expressed as follows:

\begin{equation}
\label{eq_3}
z_{v2}(t_{2})=
\begin{cases}
1, & \mbox{for } t_{2} \geq \delta_{t_{2}} \\
0, & \mbox{for } t_{2} < \delta_{t_{2}},
\end{cases}	
\end{equation}
where $z_{v2}$ represents an indicator function for the IoI state from the visual attention state. Typical values for $\delta_{t_{1}}$ and $\delta_{t_{2}}$ are in the range of $2-3 sec$.

Based on the two indicator functions described above, we can represent the indicator function for the IoI state as follows:

\begin{equation}
\label{eq_4}
z_{v}(z_{v1}(t_{1}), z_{v2}(t_{2}))=
\begin{cases}
1, & \mbox{for } z_{v1}(t_{1})=1 \mbox{ and } z_{v2}(t_{2})=1 \\
0, & \mbox{otherwise}.
\end{cases}	
\end{equation}

The final indicator function for the IoI state that integrates both the audio and vision-based IoI detection and the vision-based IoI detection is:

\begin{equation}
\label{eq_4}
z_{IoI}(z_{a}, z_{v})=
\begin{cases}
1, & \mbox{for } z_{a}=1 \mbox{ or } z_{v}=1 \\
0, & \mbox{otherwise}.
\end{cases}	
\end{equation}
From the above equation, the robot can detect the IoI and initiate a proper service protocol which depends on what the user speaks. Note that the selection problem of a proper robot service through speech recognition and analysis is beyond the scope of this paper and is not discussed.

\section{Experimental Results}
\label{sec:experiment}
We carried out experiments to verify that the proposed framework is useful for a robot to detect the IoI. In the experiments, our previous monitoring robot \cite{kwon2023heterogeneous} is used. The experiments were set up to evaluate the IoI detection performance of a robot.

\subsection{Experiment Setup}

\begin{figure}[t]
\vspace{0.3cm}
\centering
    \includegraphics[width=0.8 \linewidth]{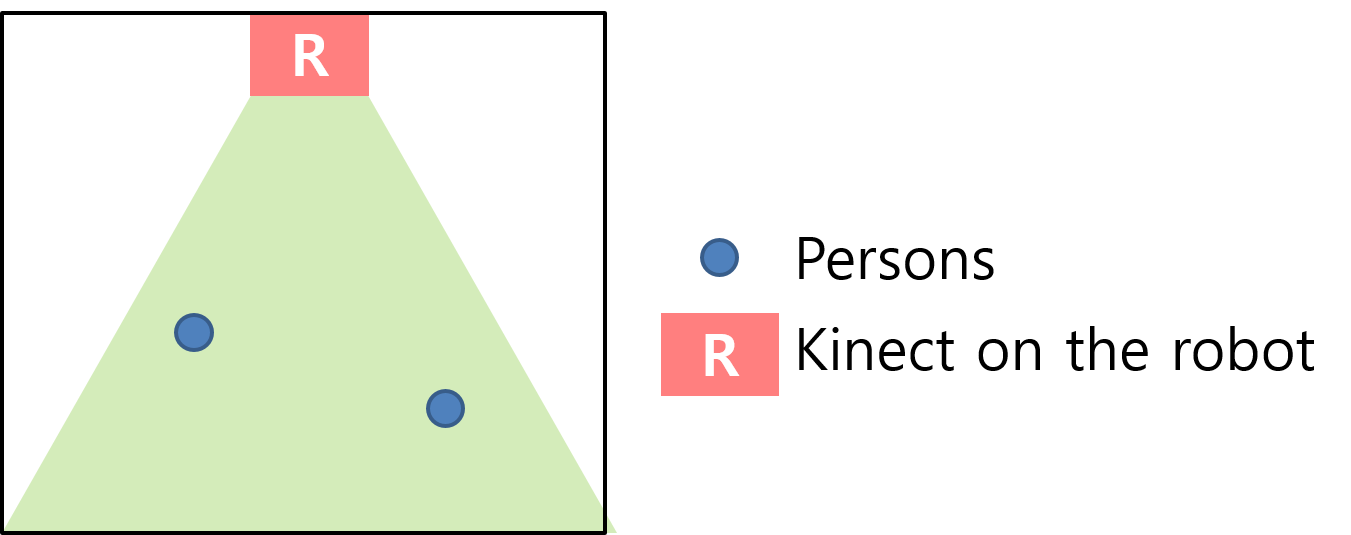}
    \caption{\textbf{An illustration of the experiment environment}.}
    \label{fig_4}
\end{figure}

The experiments are set up in a domestic environment, as shown in Fig.~\ref{fig_4}. In our experiment, we installed an Azure Kinect sensor on the robot for face direction detection as well as for person detection and tracking. More Kinect sensors can be installed in the environment with calibration to deal with larger or more complex environments. Also, we employ an audio sensor of the Kinect which involves a seven-microphone array. In the experimental environment, two participants are free to sit, stand, or walk, provided they remain within the vision sensor's field of view, which is essential for the person detection and tracking component.

\begin{figure}[t]
\vspace{0.3cm}
\centering
    \includegraphics[width=0.8 \linewidth]{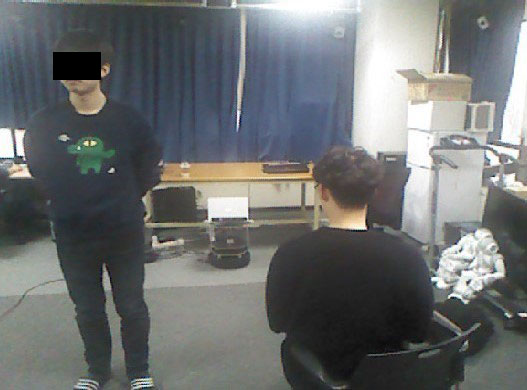}
    \caption{\textbf{An example scene from the Azure Kinect sensor on the robot}. One user, who intends to interact with the robot, is positioned facing the robot after making a sound. Conversely, the other user, who does not intend to interact, is looking away from the robot.}
    \label{fig_5}
\end{figure}

Fig.~\ref{fig_5} shows sample images obtained from the Kinect sensor. Besides typical non-IoI scenarios, the experimental data also encompass the following representative situations, in line with our IoI detection strategy:

\begin{itemize}
    \item The user speaks to the robot while looking at it.
    \item The user looks at the robot continuously.
    \item The user temporarily faces the robot while he or she is looking around.
    \item The user do not speak, but a radio is turning on.
\end{itemize}


\subsection{Performance Evaluation}

\begin{table}[t]
	\renewcommand{\arraystretch}{1.3}
	\caption{Performance comparison of the IoI detection algorithms}
	\label{table_1}
	\begin{center}
		\begin{tabular}{cccc}
			\hline
			       & Precision & Recall & F-measure \\
			\hline
			AV-IoI   & 82.35 \% & 70 \% & 75.68 \% \\
			Full-IoI & 86.36 \% & 95 \% & 90.48 \% \\
			\hline
		\end{tabular}
	\end{center}
\end{table}

To evaluate the performance of the proposed framework, we analyzed the proposed IoI detection algorithm in terms of precision, recall, and F-measure. The precision is defined as the fraction of relevant instances among the retrieved instances, which means how many detections are correct. The recall is defined as the fraction of relevant instances that have been retrieved over the total amount of relevant instances, which indicates how well the algorithm detects targets without missing. The F-measure is defined as the harmonic mean of the precision and recall. The F-measure is frequently used in the field of information retrieval for measuring classification performance, because both the precision and recall give different information that can complement each other when combined, and therefore if one of them excels more than the other, the F-measure gives a single metric that can reflect it.

Table~\ref{table_1} shows a performance comparison of the proposed audio and vision-based IoI detection (denoted as AV-IoI) and that of the proposed IoI detection which includes vision-based IoI detection as well (denoted as Full-IoI). It is shown that the proposed method can efficiently deal with a situation in which the user looks at the robot without speaking. 

The noise such as the sound of a radio or a TV substantially affects the increase of both the false positive and false negative error rates for the sound source localization component. In the proposed framework, however, we combine audio and vision sensors, and therefore we can efficiently reduce the false detections caused by the noise. Additionally, by accounting for interactions that occur without verbal utterance, this approach offers the advantage of facilitating IoI detection for users who may have difficulties speaking due to illness or other conditions.

However, the proposed IoI detection algorithm has a drawback that the detection error occurred when the user spoke in a small voice. It is not easy for the robot audio sensor to catch the user's small voice when he or she is far away from the robot. Also, the detection failed when the face detection component missed the user's frontal face.

\section{CONCLUSIONS}
\label{sec:conclusions}
In this paper, we propose a new IoI detection framework without any hotword for HRI based on audio and vision sensor fusion. In the proposed framework, the IoI is detected when the user starts to speak while looking at the robot or the user looks at the robot continuously. Thus, the proposed algorithm does not need any predefined special keyword or markers that the user should keep attaching on their own body, which can make the proposed algorithm applicable to various environments. To implement this, a state transition model for the proposed IoI detection framework is designed and verified by experiments with a mobile robot. Initially, the robot keeps monitoring state. Then, the robot's state can be changed according to the perceived audio and vision sensor data. In the proposed state transition model, the IoI state is attainable via two pathways: one involving both vocal and visual attention states, and the other exclusively through the visual attention state. To integrate our model into a robotic architecture, all components are implemented and combined within the ROS environment. The performance of the proposed IoI detection algorithm are evaluated through the experiments in a domestic environment.


\section*{ACKNOWLEDGMENT}

This work was supported by the Government-wide R\&D Fund for Infections Disease Research (GFID), funded by the Ministry of the Interior and Safety, Republic of Korea (grant number : 20014463), by the Technology Innovation Program and Industrial Strategic Technology Development Program (20018256, Development of service robot technologies for cleaning a table), and by the National Research Foundation of Korea(NRF) grant funded by the Korea government(MSIT)(No.RS-2023-00210863).

\bibliographystyle{IEEEtran}
\bibliography{IEEEabrv, egbib}

\addtolength{\textheight}{-12cm}   

\end{document}